\documentclass[11pt,a4paper]{article}
\usepackage{times,latexsym}
\usepackage{url}
\usepackage[T1]{fontenc}
\usepackage{soul}

\usepackage[acceptedWithA]{tacl2018v2}
\usepackage[]{tacl2018v2}

\usepackage{xspace,mfirstuc,tabulary}

\newif\iftaclinstructions
\taclinstructionsfalse 
\iftaclinstructions
\renewcommand{\confidential}{}
\renewcommand{\anonsubtext}{(No author info supplied here, for consistency with
TACL-submission anonymization requirements)}
\newcommand{\instr}
\fi

\iftaclpubformat 

\else

\fi

\usepackage{amsmath}
\usepackage{arydshln}
\usepackage[ruled,vlined]{algorithm2e}

\usepackage{graphicx}
\graphicspath{ {./graphics/} }

\usepackage{todonotes}

\newcommand{\atrain}{\ensuremath{\mathcal{A}_{train}}\xspace} 
\newcommand{\apin}{\ensuremath{\mathcal{A}_{in\_probe}}\xspace} 
\newcommand{\apout}{\ensuremath{\mathcal{A}_{out\_probe}}\xspace} 
\newcommand{\aood}{\ensuremath{\mathcal{A}_{ood}}\xspace} 
\newcommand{\ball}{\ensuremath{\mathcal{B}_{all}}\xspace} 
\newcommand{\btrain}[1]{\ensuremath{\mathcal{B}_{train}^{#1}}\xspace} 
\newcommand{\bpin}[1]{\ensuremath{\mathcal{B}_{in\_probe}^{#1}}\xspace} 
\newcommand{\bpout}[1]{\ensuremath{\mathcal{B}_{out\_probe}^{#1}}\xspace} 
\newcommand{\e}{\ensuremath{e_i^{(d)}}\xspace} 
\newcommand{\f}{\ensuremath{f_i^{(d)}}\xspace} 
\newcommand{\fe}{(\f,\e)\xspace} 
\newcommand{\h}{\ensuremath{\hat{e}_i^{(d)}}\xspace} 

\title{Membership Inference Attacks on Sequence-to-Sequence Models: \\
Is My Data In Your Machine Translation System?}

\author{Sorami Hisamoto\Thanks{Work done while visiting Johns Hopkins University.} \qquad Matt Post \qquad Kevin Duh \\
 Works Applications \qquad  Johns Hopkins University \\
\qquad \qquad \quad {\tt s@89.io} \qquad {\tt \{post,kevinduh\}@cs.jhu.edu} \\
}

\date{}

\begin{document}
\maketitle
\begin{abstract}
Data privacy is an important issue for ``machine learning as a service'' providers. 
We focus on the problem of membership inference attacks: given a data sample and black-box access to a model's API, determine whether the sample existed in the model's training data. 
Our contribution is an investigation of this problem in the context of sequence-to-sequence models, which are important in applications such as machine translation and video captioning.
We define the membership inference problem for sequence generation, provide an open dataset based on state-of-the-art machine translation models, and report initial results on whether these models leak private information against several kinds of membership inference attacks. 
\end{abstract}

\section{Motivation}

There are many situations where private entities are worried about the privacy of their data.
For example, many companies provide black-box training services where users are able to upload their data and have customized models built for them, without requiring machine learning expertise. 
A common concern in these ``machine learning as a service'' offerings is that the uploaded data be visible only to the client that owns it.

Currently, these entities are in the position of having to trust that service providers abide by the terms of their agreements.
While trust is an important component in relationships of all kinds, it has its limitations.
In particular, it falls short of a well known security maxim, originating in a Russian proverb that translates as, \emph{Trust, but verify}.\footnote{Popularized by Ronald Reagan in the context of nuclear disarmament.}
Ideally, customers would be able to verify that their private data was not being slurped up by the serving company, whether by design or accident.

\begin{figure}[t]
    \centering
    \includegraphics[width=0.5\textwidth]{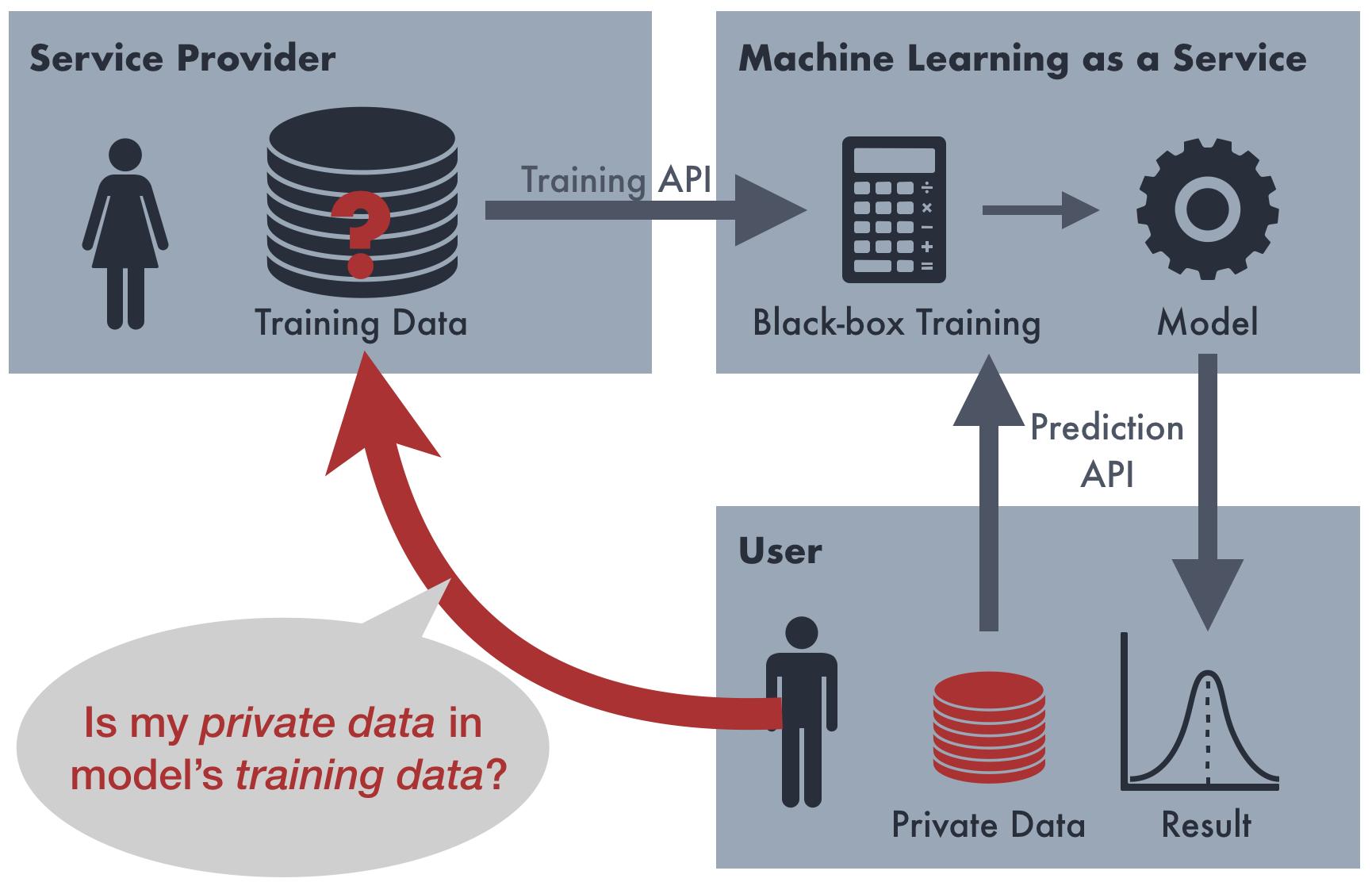}
    \caption{Membership Inference Attack}
    \label{fig:alice_and_bob}
\end{figure}

This problem has been formalized as the \textit{membership inference} problem, first introduced by \newcite{shokri2017membership} and defined as:
``Given a machine learning model and a record, determine whether this record was used as part of the model's training dataset or not.''
The problem can be tackled in an adversarial framework: the attacker is interested in answering this question with high accuracy, while the defender would like this question to be unanswerable (see Figure~\ref{fig:alice_and_bob}).
Since then, researchers have proposed many ways to attack and defend the privacy of various types of models. 
However, the work so far has only focused on standard classification problems, where the output space of the model is a fixed set of labels. 

In this paper, we propose to investigate membership inference for \textit{sequence generation} problems, where the output space can be viewed as a chained sequence of classifications.
Prime examples of sequence generation includes machine translation and text summarization: in these problems, the output is a sequence of words whose length is undetermined a priori.
Other examples include speech synthesis and video caption generation.
Sequence generation problems are more complex than classification problems, and it is unclear whether the methods and results developed for membership inference in classification problems will transfer. 
For example, one might imagine that while a flat classification model might leak private information when the output is a single label, a recurrent sequence generation model might obfuscate this leakage when labels are generated successively with complex dependencies.

We focus on machine translation (MT) as the example sequence generation problem. Recent advances in neural sequence-to-sequence models have improved the quality of MT systems significantly, and many commercial service providers are deploying these models via public API's. 
We pose the main question in the following form:
\begin{quote}
\emph{Given black-box access to an MT model, is it possible to determine whether a particular sentence pair was in the training set for that model?}
\end{quote}
%

In the following, we define membership inference for sequence generation problems (\S\ref{sec:problem}) and contrast with prior work on classification (\S\ref{sec:priorwork}). 
Next we present a novel dataset (\S\ref{sec:data}) based on state-of-the-art MT models.\footnote{We release the data to encourage further research in this new problem: \url{https://github.com/sorami/tacl-membership}} 
Finally, we propose several attack methods (\S\ref{sec:attacks}) and present a series of experiments evaluating their ability to answer the membership inference question (\S\ref{sec:results}). 
Our conclusion is that simple one-off attacks based on shadow models, which proved successful in classification problems, are not successful on sequence generation problems; this is a result that favors the defender. 
Nevertheless, we describe the specific conditions where sequence-to-sequence models still leak private information, and discuss the possibility of more powerful attacks (\S\ref{sec:discussion_conclusion}).

\section{Problem Definition}
\label{sec:problem}

We now define the membership inference attack problem for sequence-to-sequence models in detail. Following tradition in the security research literature, we introduce three characters:

\paragraph{Alice (\textit{the service provider})} builds a sequence-to-sequence model based on an undisclosed dataset \atrain and provides a public API. For MT, this API takes a foreign sentence $f$ as input and returns an English translation $\hat{e}$.

\paragraph{Bob (\textit{the attacker})} is interested in discerning whether a data sample was included in Alice's training data \atrain by exploiting Alice's API. This sample is called a ``probe'' and consists of a foreign sentence $f$ and its reference English translation, $e$. Together with the API's output $\hat{e}$, Bob has to make a binary decision using a membership inference classifier $g(\cdot)$, whose goal is to predict:\footnote{In the experiments, we will also consider extending the information available to Bob. For example, if Alice additionally provides the translation probabilities $\rho$ in the API, then Bob can exploit that in the classifier as $g(f,e,\hat{e},\rho)$.}
    \begin{equation}
         g(f,e,\hat{e}) =  
        \begin{cases}
                \textbf{in}& $if probe $ \in \atrain\\
                \textbf{out}& $otherwise$
        \end{cases}
    \end{equation}
    We term \textit{in-probes} to be those probes where the true class is \textbf{in}, and \textit{out-probes} to be those whose  true class is \textbf{out}. Importantly, note that Bob has access not only to $f$ but also to $e$ in the probe. Intuitively, if $\hat{e}$ is equivalent to $e$, then Bob may believe that the probe was contained in \atrain; however, it may also be possible that Alice's model generalizes well to new samples and translates this probe correctly. The challenge for Bob is to make this distinction; the challenge for Alice is to prevent Bob from doing so.

\paragraph{Carol (\textit{the neutral third-party})} is in charge of setting up the experiment between Alice and Bob. She decides which data samples should be used as in-probes and out-probes and evaluates Bob's classification accuracy. Carol is introduced only to clarify the exposition and to setup a fair experiment for research purposes. In practical scenarios, Carol does not exist: Bob decides his own probes, and Alice decides her own \atrain.

\subsection{Detailed Specification}
\label{subsec:probdetail}
In order to be precise about how Carol sets up the experiment, we will explain in terms of machine translation, but note that the problem definition applies to any sequence-to-sequence problem. 
A training set for MT consists of a set of sentence pairs \{\fe\}. We use a label $d \in \{\ell_1,\ell_2,\ldots\}$ to indicate the domain (the subcorpus or the data source), and an index $i \in \{1,2,\ldots,I(d)\}$ to indicate the sample id in the domain (subcorpus). 
For example, \e with $d=\ell_1$ and $i=1$ might refer to the first sentence in the \verb|Europarl| subcorpus, while \e with $d=\ell_2$ and $i=1$ might refer to the first sentence in the \verb|CommonCrawl| subcorpus. $I(d)$ is the maximum number of sentences in the subcorpus with label $d$. 
The distinction among subcorpora is not necessary in the abstract problem definition, but is important in practice when differences in data distribution may reveal signals in membership. 

Without loss of generality, in this section assume that Carol has a finite number of samples from two subcorpora $d\in\{\ell_1,\ell_2\}$.
First, she creates an out-probe of $k$ samples from subcorpus $\ell_1$: 

\begin{equation}
\apout = 
        \left\{
            \fe:
            \begin{aligned}
                &d=\ell_1,\ell_2 \\
                &i=1,\ldots,k
            \end{aligned}
        \right\}
\end{equation}

Then Carol creates the data for Alice to train Alice's MT model, using subcorpora $\ell_1$ and $\ell_2$:
\begin{equation}
\atrain = 
        \left\{
            \fe:
            \begin{aligned}
                &d=\ell_1,\ell_2 \\
                &i=k+1,\ldots,I(d) \\
            \end{aligned}
        \right\}
\end{equation}

Importantly, the two sets are totally disjoint: i.e.~$\apout \cap \atrain = \emptyset$. 
By definition, out-probes are sentence pairs that are not in Alice's training data.
Finally, Carol creates the in-probe of $k$ samples by drawing from \atrain, i.e.~$\apin \subset \atrain$, which is defined to be samples that are included in training:

\begin{equation}
\apin = 
        \left\{
            \fe:
            \begin{aligned}
                &d=\ell_1,\ell_2 \\
                &i=k+1,\ldots,2k
            \end{aligned}
        \right\}
\end{equation}

Note that both \apin and \apout are sentence pairs that come from the same subcorpus; the only difference is that the former is included in \atrain while the latter is not. 

There are several ways in which Bob's data can be created. For this work, we will assume that Bob also has some data to train MT models, in order to mimic Alice and design his attacks.
This data could either be disjoint from \atrain, or contain parts of \atrain. We choose the latter, which assumes that there might be some public data that is accessible to both Alice and Bob.
This scenario slightly favors Bob. 
In the case of MT, parallel data can be hard to come by, and datasets like \texttt{Europarl} are widely accessible to anyone, so presumably both Alice and Bob would use it. However, we expect that Alice has in-house dataset (e.g., crawled data) which Bob does not have access to.
Thus, Carol creates data for Bob:

\begin{equation}
\ball = 
        \left\{
            \fe:
            \begin{aligned}
                &d=\ell_1 \\
                &i=2k+1,\ldots,I(d) \\
            \end{aligned}
        \right\}
\end{equation}

Note that this dataset is like \atrain but with two exceptions: all samples from subcorpora $\ell_2$ and all samples from \apin are discarded. One can view $\ell_2$ as Alice's own in-house corpus which Bob has no knowledge of or access to, and $\ell_1$ as the shared corpus where membership inference attacks are performed. 

To summarize, Carol gives \atrain to Alice, who uses it in whatever way she chooses to build a sequence-to-sequence model $M[\atrain, \Theta]$. The model is trained on \atrain with hyperparameters $\Theta$ (e.g., neural network architecture) known only to Alice. In parallel, Carol gives $\mathcal{B}_{all}$ to Bob, who uses it to design various attack strategies, resulting in a classifier $g(\cdot)$ (see Section \ref{sec:attacks}). When it is time for evaluation, Carol provides both probes \apin and \apout to Bob in randomized order and asks Bob to classify each sample as \textbf{in} or \textbf{out}.
For each probe \fe, Bob is allowed to make one call to Alice's API to obtain \h.

As an additional evaluation, Carol creates a third probe based on a new subcorpus $\ell_3$. We call this the ``out-of-domain (OOD) probe'':

\begin{equation}
\aood = 
        \left\{
            \fe:
            \begin{aligned}
                &d=\ell_3 \\
                &i=1,\ldots,k
            \end{aligned}
        \right\}
\end{equation}

Both \apout and \aood should be classified as \textbf{out} by Bob's classifier. However, it has been known that sequence-to-sequence models behave very differently on data from domains/genre that is significantly different from the training data \cite{koehn2017six}. The goal of having two \textbf{out} probes is to quantify the difficulty or ease of membership inference in different situations. 

\begin{figure}[t]
\includegraphics[width=0.72\columnwidth]{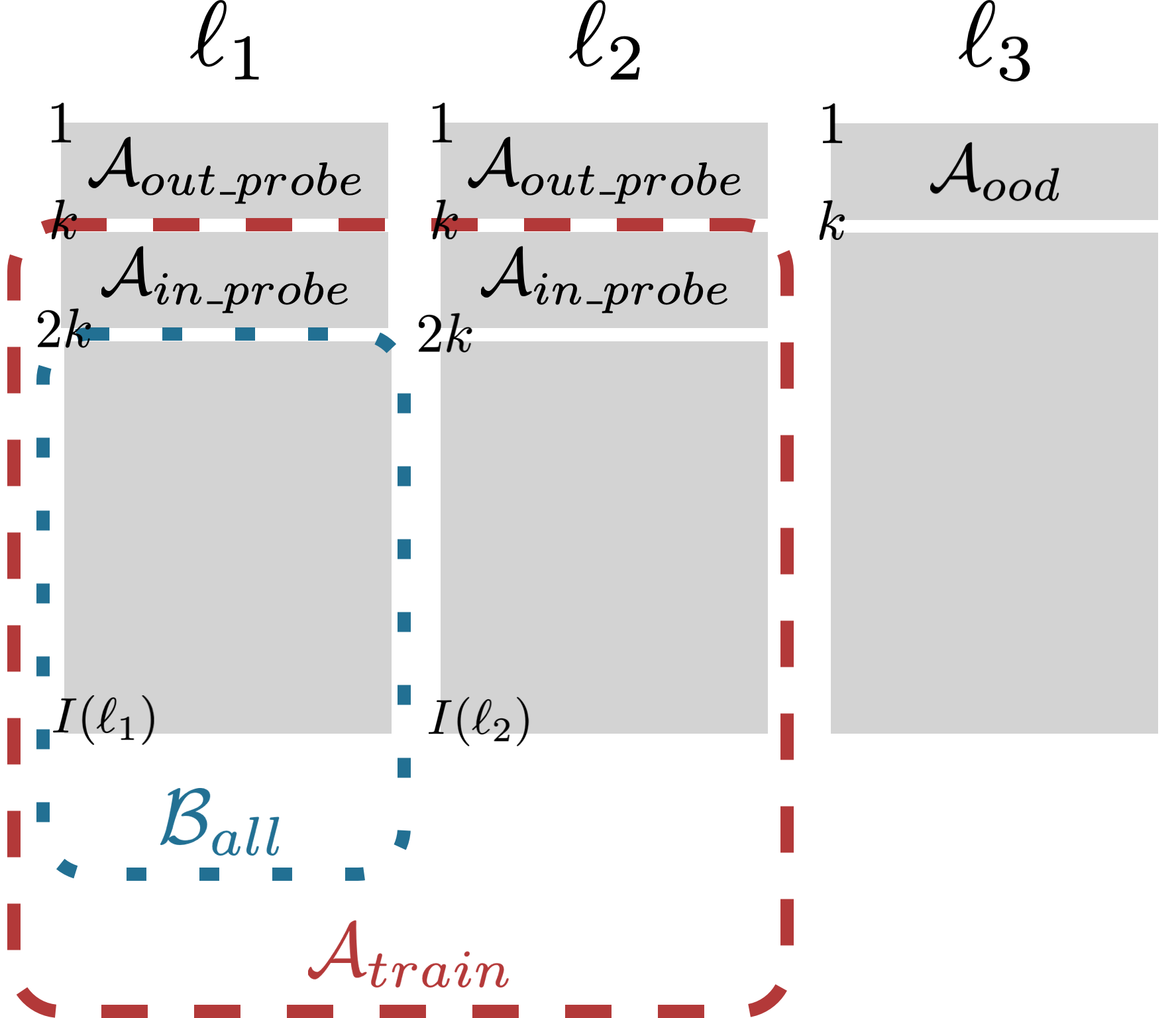}
\centering
\caption{Illustration of data splits for Alice and Bob. There are $k$ samples each for \apin, \apout, and \aood. Alice's training data \atrain excludes \apout and $\ell_{3}$, while including \apin. Bob's data \ball is a subset of Alice's data, excluding \apin and $\ell_{2}$.}
\label{fig:data}
\end{figure}

\subsection{Summary and Alternative Definitions}
\label{subsec:prob_summary}

Figure \ref{fig:data} summarizes the problem definition. The probes \apout and \aood are by construction outside of Alice's training data \atrain, while the probe \apin is included. Bob's goal is to produce a classifier that can make this distinction.
He has at his disposal a smaller dataset \ball, which he can use in whatever way he desires. 


There are alternative definitions of this membership inference problem. 
For example, one can allow Bob to make multiple API calls to Alice's model for each probe. 
This enlarges the repository of potential attack strategies for Bob. 
Or, one could evaluate Bob's accuracy not on a per-sample basis, but allow for a coarser granularity where Bob can aggregate inferences over multiple samples. 
There is also a distinction between white-box and black-box attacks: we focus on the black-box case where Bob has no internal access to the internal parameters of Alice's model, but can only guess at likely model architectures. 
In the white-box case, Bob would have access to Alice's model internals, so different attacks would be possible (e.g., backpropagation of gradients). 
In these respects, our problem definition makes the problem more challenging for Bob the attacker. 

Finally, note that Bob is not necessarily always the ``bad guy''. 
Some examples of who Alice and Bob might be in MT are:
(1) Organizations (Bob) that provide bitext data under license restrictions might be interested to determine whether their licenses are being complied with in published models (Alice).  
(2) The organizers (Bob) of an annual bakeoff, e.g.~WMT,  might wish to confirm that the participants (Alice) are following the rules of not training on test data.
(3) ``MT as a service'' providers may support customized engines if users upload their own bitext training data. The provider promises that the user-supplied data will not be used in the customized engines of other users, and can play  both Alice and Bob, attacking its own model to provide guarantees to the user. 
\textcolor{black}{If it is possible to construct a successful membership inference mechanism, then many ``good guy'' would be able to provide the aforementioned fairness (1, 2) and privacy guarantees (3). 
}

\section{Related Work}
\label{sec:priorwork}

\newcite{shokri2017membership} introduced the problem of membership inference attacks on machine learning models. 
They showed that with shadow models trained on either realistic or synthetic datasets, Bob can build classifiers that can discriminate \apin and \apout with high accuracy.
They focus on classification problems such as CIFAR image recognition and demonstrate successful attacks on both convolutional neural net models as well as the models provided by Amazon ML. 

Why do these attacks work? 
The main information exploited by Bob's classifier is the output distribution of class labels returned by Alice's API. 
The prediction uncertainty differs for data samples inside and outside the model training data, and this can be exploited. 
\newcite{shokri2017membership} proposes defense strategies for Alice, such as restricting the prediction vector to top-$k$ classes, coarsening the values of the output probabilities, and increasing the entropy of the prediction vector.
The crucial difference between their work and ours, besides our focus on sequence generation problems, is the availability of this kind of output distribution provided by Alice. 
While it is common to provide the whole distribution of output probabilities in classification problems, this is not possible in sequence generation problems because the output space of sequences is exponential in the output length.
At most, sequence models can provide a score for the output prediction \h, for example with a beam search procedure, but this is only one number and not normalized. 
We do experiment with having Bob exploit this score (Table~\ref{table:result_mt_score}), but it appears far inferior to the use of the whole distribution available in classification problems. 

Subsequent work on membership inference has focused on different angles of the problem.
\newcite{salem2018MLLeaks} investigated the effect of training the shadow model and datasets that match or does not match the distribution of \atrain, and compared training a single shadow model as opposed to many. 
\newcite{truex2018demystify} presents a comprehensive evaluation of different model types, training data, and attack strategies;
Borrowing ideas from adversarial learning and minimax games, 
\newcite{hayes2017gan} proposes attack methods based on generative adversarial networks, while \newcite{nasr2018regularization} provides adversarial regularization techniques for the defender.
\newcite{nasr2019whitebox} extends the analysis to white-box attacks and a federated learning setting.
\newcite{pyrgelis2018location} provides an empirical study on location data.
\newcite{veale2018law} discusses membership inference and the related model inversion problem, in the context of data protection laws like GDPR.

\citet{shokri2017membership} notes a synergistic connection between the goals of learning and the goals of privacy in the case of membership inference: the goal of learning is to generalize to data outside the training set (e.g., so that \apout and \aood are translated well), while the goal of privacy is to prevent leaking information about data in the training set. The common enemy of both goals is overfitting. 
\newcite{yeom2017overfitting} analyze how  overfitting by Alice's increases the risk  privacy leakage; \newcite{long2018understanding} showed that even a well-generalized model holds such risks in classification problems, implying that overfitting by Alice is a sufficient but not necessary condition for privacy leakage. 

A large body of work exists in differential privacy \cite{dwork2008survey,machanavajjhala2017tutorial}.
Differential privacy provides guarantees that a model trained on some dataset \atrain will produce statistically similar predictions as a model trained on another dataset which differs in exactly one sample.  
This is one way in which Alice can defend her model \cite{rahman2018differential}, but note that differential privacy is a stronger notion and often involves a cost in Alice's model accuracy.
Membership inference assumes that content of the data is known to Bob and only is concerned whether it was used. Differential privacy also protects the content of the data (i.e., the actual words in \fe should not be inferred).


\textcolor{black}{
\newcite{Song:2019:ADP:3292500.3330885} explored the membership inference problem of natural language text, including word prediction and dialog generation. They assume that the attacker has access to a probability distribution or a sequence of distributions over the vocabulary for the generated word or sequence. This is different from our work where the attacker gets only the output sequence, which we believe is a more realistic setting.}

\section{Data and Evaluation Protocol}
\label{sec:data}

\subsection{Data: subcorpora and splits}

Based on the problem definition in Section \ref{sec:problem}, we construct a dataset to investigate the possibility of the membership inference attack on MT models. We make this dataset available to the public to encourage further research.\footnote{\url{https://github.com/sorami/tacl-membership}}

There are various considerations to ensure the benchmark is fair for both Alice and Bob:
we need a dataset that is large and diverse to ensure Alice can train state-of-the-art MT models and Bob can test on probes from different domains. 
We used corpora from the Conference on Machine Translation (WMT18) \cite{W18-6401}. 
We chose German--English language pair because it has a reasonably large amount of training data, and previous work demonstrate high BLEU scores.

We now describe how Carol prepares the data for Alice and Bob.
First, Carol selects 4 subcorpora for the training data of Alice, namely \verb|CommonCrawl|, \verb|Europarl v7|, \verb|News Commentary v13|, and \verb|Rapid 2016|. A subset of these 4 subcorpora are also available to Bob ($\ell_1$ in section \ref{subsec:probdetail}). 
In addition, Carol gives \verb|ParaCrawl| to Alice but not Bob ($\ell_2$ in \S\ref{subsec:probdetail}). We can think of it as an in-house data the service provider holds.
For all these subcorpora, Carol first performs basic preprocessing: (a) tokenization of both the German and English sides using the Moses tokenizer, (b) de-duplication of sentence pairs so that only unique pairs are present, and (c) randomly shuffling all sentences prior to splitting into probes and MT training data.\footnote{These are design decisions that balance between simple experimentation vs. realistic condition. Carol doing a common tokenization removes some of the MT-specific complexity for researchers who want to focus on the Alice or Bob models. However, in a real-world public API, Alice's tokenization is likely to be unknown to Bob. We decided on a middle ground to have Carol perform a common tokenization, but Alice and Bob do their own subword segmentation.
}

\begin{figure}[t]
\includegraphics[width=0.9\columnwidth]{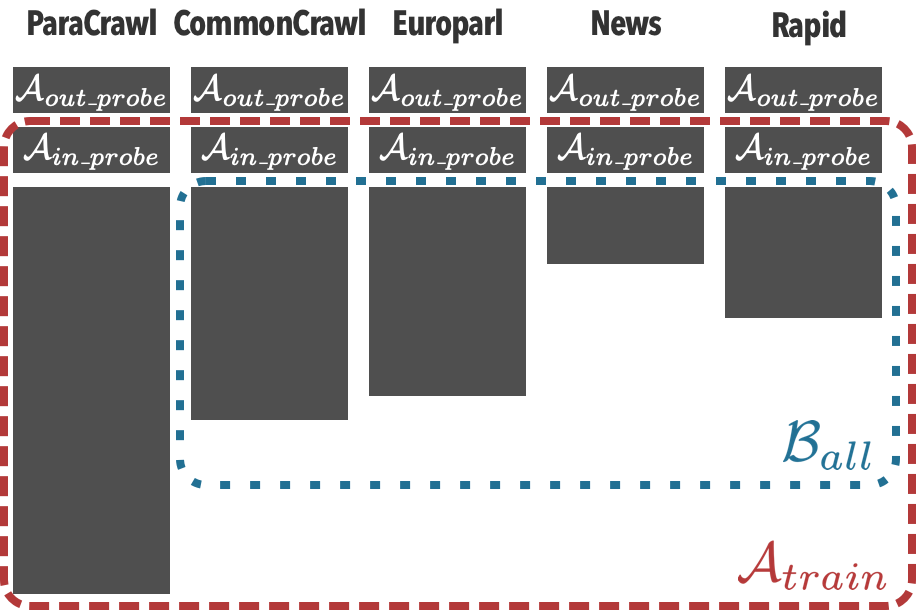}
\centering
\caption{Illustration of actual MT data splits.
\atrain does not contain \apout, and \ball is a subset of \atrain with \apin and \texttt{ParaCrawl} excluded.
}
\label{fig:actual_data}
\end{figure}

\begin{table*}[t]
\centering
\begin{tabular}{ c c c c c c } 
 \hline
  & \apout & \apin & \atrain & \ball & \aood \\ 
 \hline
\verb|ParaCrawl| & 5,000 & 5,000 & 4,518,029 & 0 & N/A \\ 
\verb|CommonCrawl| & 5,000 & 5,000 & 2,389,123 & 2,379,123 & N/A \\
\verb|Europarl| & 5,000 & 5,000 & 1,865,271 & 1,855,271 & N/A \\ 
\verb|News| & 5,000 & 5,000 & 273,702 & 263,702 & N/A \\ 
\verb|Rapid| & 5,000 & 5,000 & 1,062,214 & 1,052,214 & N/A \\ 
\hdashline
\verb|EMEA| & N/A & N/A & N/A & N/A & 5,000 \\ 
\verb|Koran| & N/A & N/A & N/A & N/A & 5,000 \\ 
\verb|Subtitles| & N/A & N/A & N/A & N/A & 5,000 \\ 
\verb|TED| & N/A & N/A & N/A & N/A & 5,000 \\ 
\hline
TOTAL & 25,000 & 25,000 & 10,108,339 & 5,550,310 & 20,000 \\
 \hline
\end{tabular}
\caption{Number of sentences per set and subcorpus. For each subcorpus, \atrain includes \apin and does not include \apout. \ball is a subset of \atrain, excluding \apin and \texttt{ParaCrawl}. \aood is for evaluation only, and only Carol has access to them.}
\label{table:datasummary}
\end{table*}

Figure~\ref{fig:actual_data} illustrates how Carol splits subcorpora for Alice and Bob.
For each subcorpus, Carol splits them to create probes \apin and \apout, and \atrain and \ball. Carol sets $k=5,000$, meaning each probe set per subcorpus has 5,000 samples. For each subcorpus, Carol selects 5,000 samples to create \apout. She then uses the rest as \atrain and select 5,000 from it as \apin. She excludes \apin and \verb|ParaCrawl| from \atrain to create a dataset for Bob, \ball.\footnote{We prepared two different pairs of \apin and \apout. Thus \ball has 10k less samples than \atrain, and not 5k less. For the experiment we used only one pair, and kept the other for future use.}
In addition, Carol has 4 other domains to create out-of-domain probe set \aood, namely, \verb|EMEA| and \verb|Subtitles 18| \cite{TIEDEMANN12.463}, \verb|Koran| (Tanzil), and \verb|TED| \cite{duh18multitarget}. These subcorpora are equivalent to $\ell_3$ in section \ref{subsec:probdetail}. The size of \aood is 5,000 per subcorpus, same as \apin and \apout.
The number of samples for each set is summarized in Table \ref{table:datasummary}.


\subsection{Alice MT Architecture}
\label{subsec:alicemt}

Alice uses her dataset \atrain (consisting of 4 subcorpora and \verb|ParaCrawl|) to train her own MT model. 
Since \verb|Paracrawl| is noisy, Alice first applied dual conditional cross-entropy filtering \cite{junczys-dowmunt2018dual}, retaining the top 4.5 million lines.
Alice then trained a joint BPE subword model \cite{sennrich2016neural} using 32,000 merge operations.
No recasing was applied.

Alice's model is a six-layer Transformer \cite{transformer} using default parameters in Sockeye \cite{sockeye}.\footnote{Three-way tied embeddings, model and embedding size 512, eight attention heads, 2,048 hidden states in the feed forward layers, layer normalization applied before each self-attention layer, and dropout and residual connections applied afterward, word-based batch size of 4,096.}
The model was trained until perplexity on \verb|newstest2017| \cite{bojar2017findings} had not improved for five consecutive checkpoints, computed every 5,000 batches.

The BLEU score \cite{P02-1040} on \verb|newstest2018| was 42.6, computed using sacreBLEU \cite{post2018call} with the default settings.\footnote{Version 1.2.12, case-sensitive, ``13a'' tokenization for comparability with WMT.}

\subsection{Evaluation Protocol}
\label{subsec:protocol}


To evaluate membership inference attacks on Alice's MT models, we use the following procedure: First, Bob asks Alice to translate $f$. Alice returns her result $\hat{e}$ to Bob. Bob also has access to the reference $e$ and use his classifier $g(f,e,\hat{e})$ to infer whether $(e,f)$ was in Alice's training data. The classification is reported to Carol, who computes ``attack accuracy''.
Given a probe set $P$ containing a list of $(f, e, \hat{e}, l)$, where $l$ is the label (\textbf{in} or \textbf{out}), this accuracy is defined as:

\begin{equation}
    accuracy(g, P) = \frac{1}{\vert P \vert}  \sum^{P} [ g(f, e, \hat{e}) = l ]
\end{equation}

If the accuracy is 50\%, then the binary classification is same as random, and Alice is safe. 
An accuracy slightly above 50\% can be considered a potential breach of privacy.

\section{Membership Inference Attacks}
\label{sec:attacks}

\subsection{Shadow Model Framework}
\label{subsec:attacks_shadow}

Bob's initial approach for attack is to use ``shadow models'', similar to \newcite{shokri2017membership}. The idea is that Bob creates MT models with his data to mimic (shadow) the behavior of Alice's MT model, then train a membership inference classifier on these shadow models.
To do so, Bob splits his data \ball into his own version of in-probe, out-probe, and training set in multiple ways to train MT models. Then he translates these probe sentences with his own shadow MT models, and use the resulting $(f, e, \hat{e})$ with its \textbf{in} or \textbf{out} label to train a binary classifier $g(f, e, \hat{e})$.
If Bob's shadow models are sufficiently similar to Alice's in behavior, this attack can work.

\begin{figure}[h]
\includegraphics[width=0.9\columnwidth]{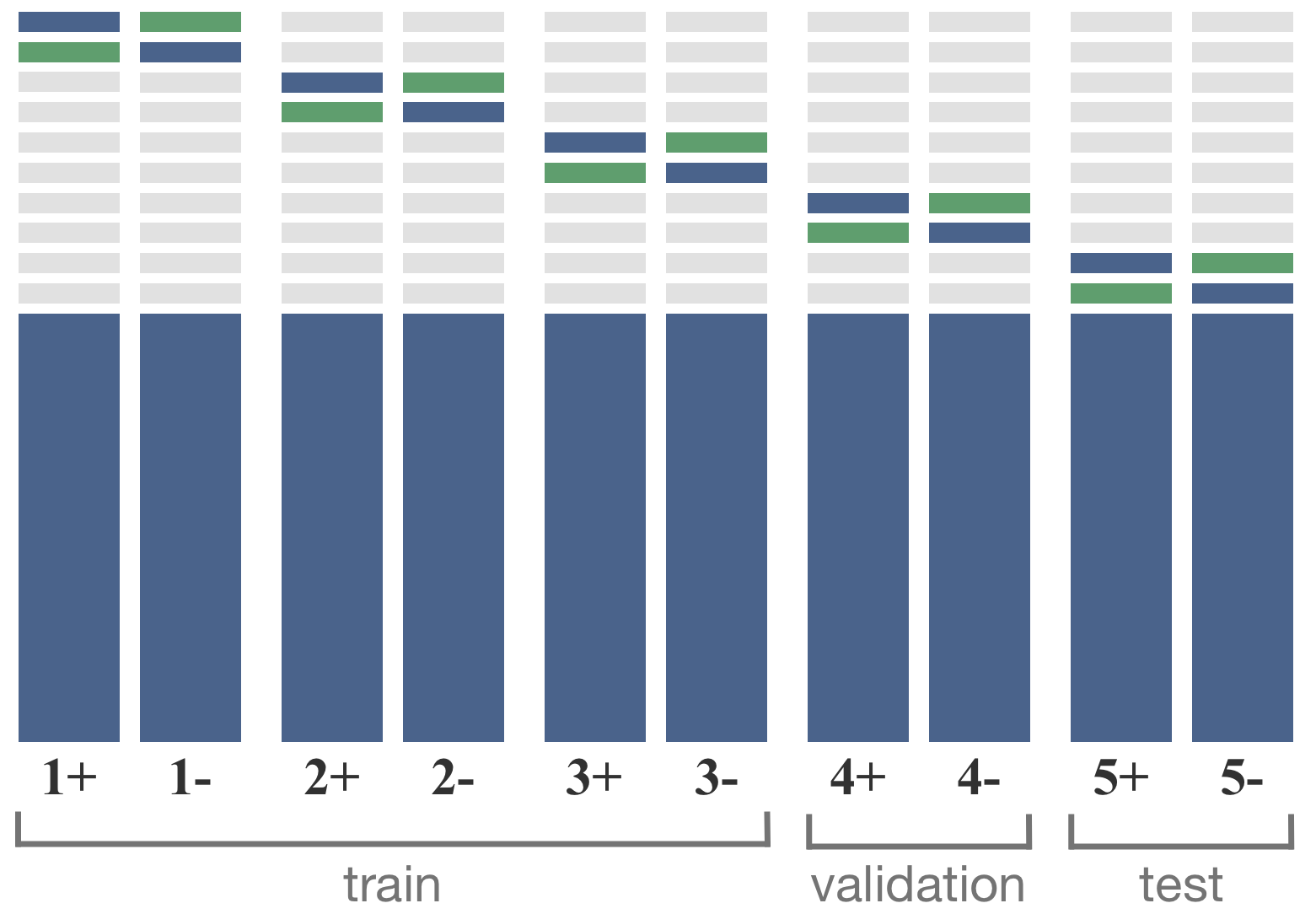}
\centering
\caption{Illustration of how Bob splits \ball for each shadow model. Blue boxes are the in-probe \bpin{} and training data \btrain{}, where small box is the in-probe and small and large boxes combined is the training data. Green box indicates the out-probe \bpout{}. Bob uses models from splits 1 to 3 as a train, 4 as a validation, and 5 as a test sets for his attack.}
\label{fig:shadowsplit}
\end{figure}

Bob first selects 10 sets of 5,000 sentences per subcorpus in \ball. He then chooses 2 sets and use one as in-probe and the other as out-probe, and combine in-probe and the rest (\ball minus 10 sets) as a training set. We use notations \bpin{1+} \bpout{1+}, and \btrain{1+} for the first group of in-probe, out-probe, and training set. Bob then swaps the in-probe and out-probe to create another group. We notate this as \bpin{1-}, \bpout{1-}, and \btrain{1-}. With 10 sets of 5,000 sentences, Bob can create 10 different groups of in-probe, out-probe, and training set. Figure~\ref{fig:shadowsplit} illustrates the data splits.

For each group of data, Bob first trains a shadow MT model using the training set. 
He then uses this model to translate sentences in the in-probe and out-probe sets. 
Bob has now a list of $(f, e, \hat{e})$ from different shadow models, and he knows for each sample if it was \textbf{in} or \textbf{out} of the training data for the MT model used to translate that sentence.

\subsection{Bob MT Architecture}


Bob's model is a 4-layer Transformer, with no tied embedding, model/embedding size 512, 8 attention heads, 1,024 hidden states in the feed forward layers, word-based batch size of 4,096. The model is optimized with Adam \cite{Kingma2015AdamAM}, regularized with label smoothing (0.1), and trained until perplexity on \verb|newstest2016| \cite{W16-2301} had not improved for sixteen consecutive checkpoints, computed every 4,000 batches. Bob has BPE subword models with vocab size 30k for each language. 
The mean BLEU scores of the ten shadow models on \verb|newstest2018| is $38.6{\pm}0.2$ (compared to 42.6 for Alice).

\subsection{Membership Inference Classifier}
\label{subsec:attacks_classifier}

Bob extracts features from $(f, e, \hat{e})$ for a binary classifier. He uses modified 1-4 gram precisions and smoothed sentence-level BLEU score \cite{P04-1077} as features. 
Bob's intuition is that if an unusually large number of n-grams in $\hat{e}$ matches $e$, then it could be a sign that this was in the training data and Alice memorized it.
Bob calculates n-gram precision by counting the number of n-grams in translation that appear in the reference sentence. 
In the later investigation Bob also considered the MT model score as an extra feature.

Bob tried different types of classifiers, namely namely Perceptron (P), Decision Tree (DT),
Na\"ive Bayes (NB), Nearest Neighbors (NN), and Multi-layer Perceptron (MLP). DT uses GINI impurity for the splitting metrics, and the max depth to be 5. 
\textcolor{black}{Our NB uses Gaussian distribution. For NN we set the number of neighbors to be 5 and used Minkowski distance.}
For MLP, we set the size of hidden layer to be 100, activation function to be ReLU, and L2 regularization term $\alpha$ to be $0.0001$.

Pseudocode~\ref{algo:classifier} summarizes the procedure to construct a membership inference classifier $g(\cdot)$ using Bob's dataset \ball.
For training the binary classifiers, Bob uses models from data splits 1 to 3 for training, 4 for validation, and 5 for his own internal testing. 
Note that the final evaluation of the attack is done using the translations of \apin and \apout with Alice MT model, by Carol.

\begin{algorithm}[]
\SetAlgoLined
\KwData{\ball}
\KwResult{$g(\cdot)$}

Split \ball into multiples groups of (\bpin{i}, \bpout{i}, \btrain{i}) \;

\ForEach{$i$ in $1+,1-,2+,2-,3+,3-$} {
  Train a shadow model $M_{i}$ using \btrain{i} \;
  Translate \bpin{i}, \bpout{i} with $M_{i}$ \;
}

Use \bpin{i}, \bpout{i}, and their translations to train $g(\cdot)$ \;

\caption{Construction of A Membership Inference Classifier}
\label{algo:classifier}
\end{algorithm}

\section{Attack Results}
\label{sec:results}

We now present a series of results based on the shadow model attack method described in Section~\ref{sec:attacks}. 
In Section \ref{subsec:mainresult} we will observe that Bob has difficulty attacking Alice under our definition of membership inference. In Sections \ref{subsec:result_ood} and \ref{subsec:result_oov} we will see that Alice nevertheless does leak some private information under more nuanced conditions. 
Section \ref{subsec:grouping} describes the possibility of attacks beyond sentence-level membership.
\textcolor{black}{Section \ref{subsec:externalresource} explore the attacks using external resources.}

\subsection{Main Result}
\label{subsec:mainresult}

\begin{table}[h]
\centering
\setlength\tabcolsep{5pt}
\begin{tabular}{ l | c : c c c } 
 \hline
  & Alice & Bob:train & Bob:valid & Bob:test  \\ 
 \hline
P & 50.0 & 50.0 & 50.0 & 50.0 \\
DT & 50.4 & 51.4 & 51.2 & 51.1 \\
NB & 50.4 & 51.2 & 51.1 & 51.0 \\
NN & 49.9 & 61.6 & 50.5 & 50.0 \\
MLP & 50.2 & 50.8 & 50.8 & 50.8 \\
 \hline
\end{tabular}
\caption{Accuracy of membership inference per classifier type, Perceptron (P), Decision Tree (DT), Na\"ive Bayes (NB), Nearest Neighbors (NN), and Multi-layer Perceptron (MLP). \textit{Alice} column shows the accuracy of attack on Alice probes \apin and \apout. \textit{Bob} columns show the accuracy on the classifiers' train, validation, and test set. Note that, following the evaluation protocol explained in \ref{subsec:protocol}, only Carol the evaluator can observe the accuracy of the attacks on Alice model.}
\label{table:result_basic}
\end{table}

\begin{figure*}[t]
    \centering
    \includegraphics[width=\textwidth]{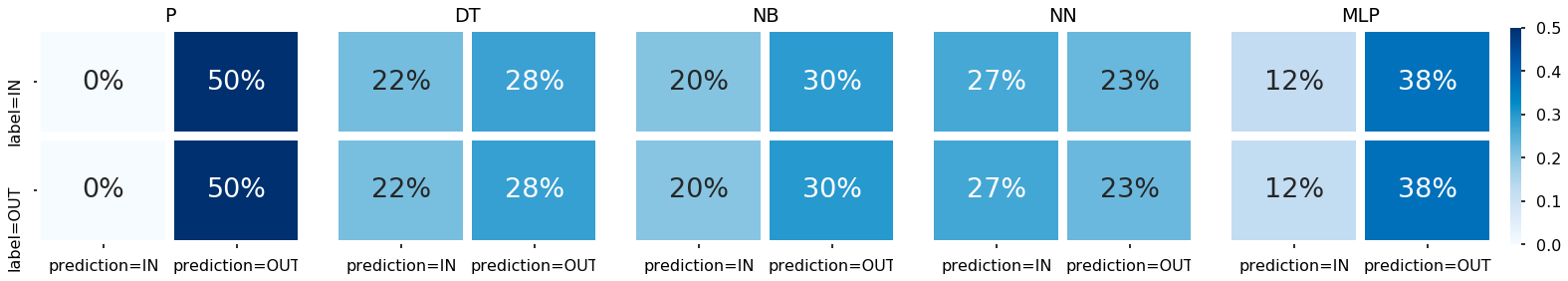}
    \caption{Confusion matrices of the attacks on Alice model per classifier type.}
    \label{fig:heatmap}
\end{figure*}

Table~\ref{table:result_basic} shows the accuracy of the membership inference classifiers. There are 5 different types of classifiers, as described in section~\ref{subsec:attacks_classifier}.
The numbers in the \textit{Alice} column shows the attack accuracy on Alice probes \apin and \apout; these are the main results.
The numbers in \textit{Bob} columns show the results on the Bob classifiers' train, validation, and test sets, as described in Section~\ref{subsec:attacks_classifier}. 

The results of the attacks on the Alice model show that it is around 50\%, meaning that the attack is not successful and the binary classification is almost the same as a random choice.
\footnote{\textcolor{black}{Some numbers are slightly over 50\% which may be interpreted as small leak of privacy. While the desired accuracy levels depend on the application, for the MT scenarios described in Section \ref{subsec:prob_summary} Bob would need much higher accuracies. For example, if Bob is a bakeoff organizer, he might want accuracy above 60\% in order to determine whether to manually check the submission. However, if Bob is providing ``MT as a service'' with strong privacy guarantees, he may need to provide the client with accuracy higher than 90\%.}}
The accuracy is around 50\% for
\textcolor{black}{Bob:valid}, meaning that Bob also has difficulty attacking his own simulated probes, therefore the poor performance on \apin and \apout is not due to  mismatches between Alice's model and Bob's model. 

The accuracy is around 50\% for Bob:train as well, reveals that the classifier $g(\cdot)$ is underfitting.\footnote{The higher accuracy for $k$-NN is an exception, but is due to having the exact same datapoint in the model as the input, which always becomes the nearest neighbor. When the $k$ value is increased, the accuracy on in-sample data decreased.} This suggests that the current features do not provide enough information to distinguish in-probe and out-probe sentences.
Figure~\ref{fig:heatmap} shows the confusion matrices of the classifier output on Alice probes. 
We see that for all classifiers, whatever prediction they make is incorrect half of the time.

\begin{table}[h]
\centering
\setlength\tabcolsep{5pt}
\begin{tabular}{ l | c : c c c } 
 \hline
  & Alice & Bob:train & Bob:valid & Bob:test  \\ 
 \hline
P & 49.7 & 49.2 & 49.3 & 49.4 \\
DT & 50.4 & 51.5 & 51.1 & 51.2 \\
NB & 50.1 & 50.2 & 50.1 & 50.2 \\
NN & 50.2 & 67.1 & 50.2 & 50.0 \\
MLP & 50.4 & 51.2 & 51.2 & 51.1 \\
 \hline
\end{tabular}
\caption{Membership inference accuracy when MT model score is added as an extra classifier feature.}
\label{table:result_mt_score}
\end{table}

Table~\ref{table:result_mt_score} shows the result when \textit{MT model score} is added as an extra feature for classification. The result indicates that this extra information does not improve the attack accuracy.
In summary, these results suggest that Bob is not able to reveal membership information at the sentence/sample level. 
This result is in contrast to previous work on membership inference in ``classification'' problems, which demonstrated high accuracy with Bob's shadow model attack.

\textcolor{black}{Additionally, note that while accuracies are close to 50\%, the number of Bob:test tend to be slightly higher than Alice's for some classifiers. This may reflect the fact that Bob:test is a matched condition using the same shadow MT architecture, while Alice probes are from a mismatched condition using an unknown MT architecture. It is important to compare both numbers in the experiments: accuracy on Alice probes is the real evaluation and accuracy on Bob:test is a diagnostic.}

\subsection{Out-of-Domain Subcorpora}
\label{subsec:result_ood}

\begin{table*}[h]
\centering
\setlength\tabcolsep{2pt}
\begin{tabular}{ l | c c c c c : c c c c } 
 \hline
 & \verb|ParaCrawl| & \verb|CommonCrawl| & \verb|Europarl| & \verb|News| & \verb|Rapid| & \verb|EMEA| & \verb|Koran| & \verb|Subtitles| & \verb|TED| \\ 
 \hline
P	& 50.0 & 50.0 & 50.0 & 50.0 & 50.0 & 100.0 & 100.0 & 100.0 & 100.0 \\
DT	& 50.3 & 51.1 & 49.7 & 50.7 & 50.0 & 67.2 & 94.1 & 80.2 & 67.1 \\
NB	& 50.1 & 51.2 & 49.9 & 50.6 & 50.2 & 69.5 & 96.1 & 81.7 & 70.5 \\
NN	& 49.4 & 50.7 & 50.3 & 49.7 & 49.2 & 43.3 & 52.6 & 48.7 & 49.9 \\
MLP	& 49.6 & 50.8 & 49.9 & 50.3 & 50.7 & 73.6 & 97.9 & 84.8 & 85.0 \\
 \hline
\end{tabular}
\caption{Membership inference accuracy per subcorpus. Right 4 columns are results for out-of-domain subcorpora. Note that \texttt{ParaCrawl} is \textit{out-of-domain} for Bob and his classifier, whereas \textit{in-domain} for Alice and her MT model.}
\label{table:result_ood}
\end{table*}

\begin{figure*}[h]
    \centering
    \includegraphics[width=0.8\textwidth]{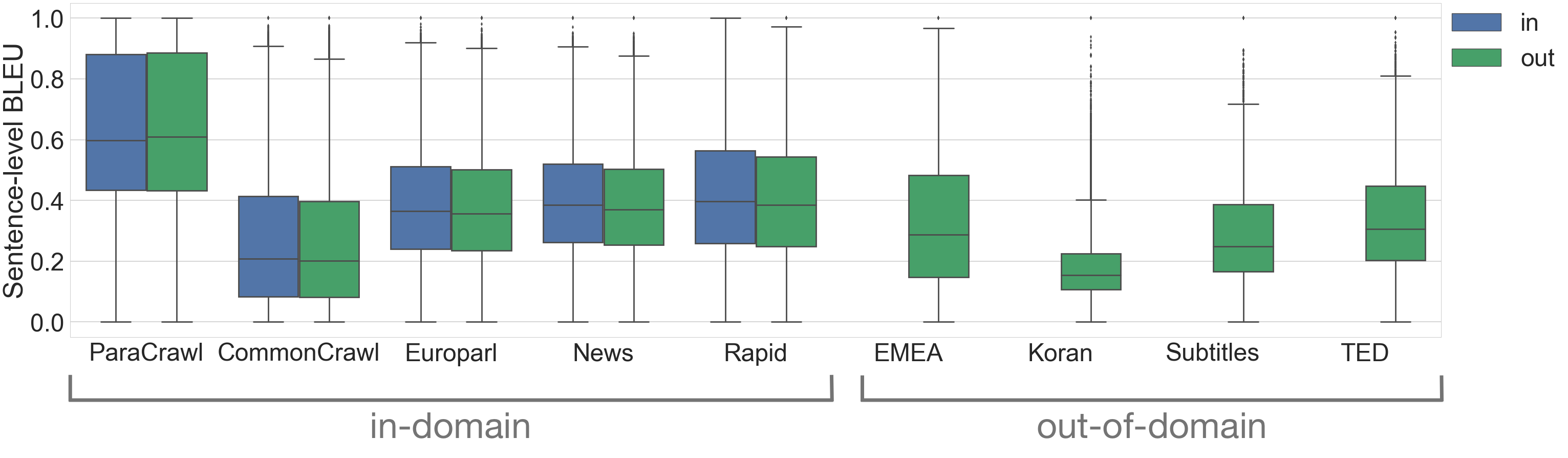}
    \caption{Distribution of sentence-level BLEU per subcorpora for \apin (blue boxes), \apout (green, left five boxes), and \aood (green, right four boxes).}
    \label{fig:boxplot}
\end{figure*}

Carol prepared out-of-domain (OOD) subcorpora, \aood, that are separate from \atrain and \ball. The membership inference accuracy of each subcorpus is shown in Table~\ref{table:result_ood}. The accuracy for OOD subcorpora are much higher than that of original in-domain subcorpora. For example, the accuracy with Decision Tree was 50.3\% and 51.1\% for \verb|ParaCrawl| and \verb|CommonCrawl| (in-domain), whereas 67.2\% and 94.1\% for \verb|EMEA| and \verb|Koran| (out-of-domain). This suggests that for OOD data Bob has a better chance to infer the membership.

\textcolor{black}{In Table~\ref{table:result_ood} we can see that Perceptron has accuracy 50\% for all in-domain subcorpora and 100\% for all OOD subcorpora. Note that the OOD subcorpora only have \textit{out-probes}; By definition none of the samples from OOD subcorpora are in the training data. We get such accuracy because our Perceptron is always predicting \textbf{out}, as we can see in Figure~\ref{fig:heatmap}. We believe this behavior is caused by applying Perceptron to inseparable data, and this particular model happened to be trained to act this way. To confirm this we have trained variations of Perceptrons by shuffling the training data, and observed that the resulting models had different output ratio of \textbf{in} and \textbf{out}, and in some cases always predicting \textbf{in} for both in and OOD subcorpora.}

Figure~\ref{fig:boxplot} shows the distribution of sentence-level BLEU scores per subcorpus. 
The BLEU scores tends to be lower for OOD subcorpora, and the classifier may exploit this information to distinguish the membership better. 
But note that \verb|EMEA| (out-of-domain) and \verb|CommonCrawl| (in-domain) have similar BLEU, but vastly different membership accuracies, so the classifier may also be exploiting n-gram match distributions. 

Overall, these results suggest that Bob's accuracy depends on the specific type of probe being tested. If there is a wide distribution of domains, there is a higher chance that Bob may be able to reveal membership information. 
\textcolor{black}{Note that in the actual scenario Bob will have no way of knowing what is OOD for Alice, so there is no signal that is exploitable for Bob. This section is meant as an error analysis that describes how membership inference classifiers behave differently in case the probe is OOD.}

\subsection{Out-of-Vocabulary Words}
\label{subsec:result_oov}

\begin{table}[h]
\centering
\setlength\tabcolsep{5pt}
\begin{tabular}{ l | c c c } 
 \hline
  & OOV in src & OOV in ref & OOV in both \\ 
 \hline
P 	& 100.0 & 100.0 & 100.0 \\
DT 	& 73.9 	& 74.1 	& 68.0 \\
NB 	& 77.4 	& 77.0 	& 70.3 \\
NN 	& 49.9 	& 49.2 	& 49.3 \\
MLP 	& 89.0 	& 85.8 	& 80.4 \\
 \hline
\end{tabular}
\caption{Membership inference accuracy on the sentences \textcolor{black}{in Bob:test} containing Out-of-vocabulary (OOV) words \textcolor{black}{for the MT model used for translation.}}
\label{table:result_oov}
\end{table}

We also focused on the samples which contain the words that never appear in the training data \textcolor{black}{of the MT model used for translation}, i.e., out-of-vocabulary (OOV) words. For this analysis, we focus only on vocabulary that does not exist in the training data of Bob's shadow \textcolor{black}{MT} models, rather than Alice's, since Bob does not have access to her vocabulary. \textcolor{black}{By definition there are only \textit{out-probes} in OOV subsets.}

For Bob's shadow models, 7.4\%, 3.2\%, and 1.9\% of samples in the probe sets had one or more OOV words in source, reference, or both sentences, respectively. 
Table~\ref{table:result_oov} shows the membership inference accuracy of the OOV subsets \textcolor{black}{from Bob test set}, which is generally very high ($>$70\%). This implies that sentences with OOV words are translated idiosyncratically compared to the ones without OOV words, and classifier can exploit this.


\subsection{Alternative Evaluation: Grouping Probes}
\label{subsec:grouping}

Section \ref{subsec:mainresult} showed it is generally difficult for Bob to determine membership for the strict definition of one sentence per probe. 
What if we loosen the problem, letting the probe be a group of sentences? 




We create probes of 500 sentences each to investigate this hypothesis.
Bob randomly samples 500 sentences with the same label from Bob's training set to form a probe group. To create sufficient training data for his classifier, Bob repeats sampling and creates 6,000 groups. 
Bob uses sentence BLEU bin percentage, and corpus BLEU as features for classification. 
For each group, Bob counts the sentence BLEU for each bin. The bin size is set to 0.01.
Bob also uses all 500 translation together to calculate the group's corpus BLEU score.
Bob trains the classifiers using these features, and apply it to Bob's validation and test set, and Alice set. 
These sets are evenly split into groups of 500, not sampled as done in training.


\begin{table}[h]
\centering
\setlength\tabcolsep{5pt}
\begin{tabular}{ l | *{3}{c} : *{2}{c} } 
 \hline
 & \multicolumn{3}{c:}{Bob} & \multicolumn{2}{c}{Alice} \\
 & train & valid & test & original & adjusted \\ 
 \hline
P & 71.6 & 69.4 & 68.1 & 50.0 & 59.0 \\
DT & 70.4 & 65.6 & 64.4 & 52.0 & 61.0 \\
NB & 72.9 & 67.5 & 70.0 & 50.0 & 50.0 \\
NN & 77.4 & 66.9 & 62.5 & 51.0 & 50.0 \\
MLP & 73.0 & 68.8 & 70.0 & 50.0 & 52.0 \\
 \hline
\end{tabular}
\caption{Attack accuracy on probe groups. In addition to the \textit{original} Alice set, we have \textit{adjusted} set where the feature values are adjusted by subtracting the mean BLEU difference between Alice and Bob models.}
\label{table:grouping_probes}
\end{table}

\textcolor{black}{
Table \ref{table:grouping_probes} shows the accuracy on probe groups. 
We can see that the accuracy is much higher than 50\%, not only for Bob's training set but also for his validation and test sets. 
However, for Alice, we found that classifiers were almost always predicting \textbf{in}, resulting the accuracy to be around 50\%. 
This is due to the fact that classifiers were trained on shadow models that have lower BLEU scores than Alice.
This suggests that we need to incorporate the information about the Alice / Bob MT performance difference.
}

\textcolor{black}{
One way to adjust the difference is to directly manipulate the input feature values. We adjusted the feature values, compensating by the difference in mean BLEU scores, and accuracy on Alice probes increased to 60\% for P and DT as shown in the ``adjusted'' column of Table~\ref{table:grouping_probes}. 
If the classifier took advantage of the absolute values in its decision, the adjustment may give improvements. If that is not the case, then improvements are less likely. Before the adjustment, all classifiers were predicting everything to be \textbf{in} for Alice probes. Classifiers like NB and MLP apparently did not change how often it predicts \textbf{in} even after the normalization, whereas classifiers like P and DT did.
In a real scenario this BLEU difference can be reasonably estimated by Bob, since he can use Alice's translation API to calculate BLEU score on a held-out set, and compare it with his shadow models.
}

\begin{figure}[]
    \centering
    \includegraphics[width=\columnwidth]{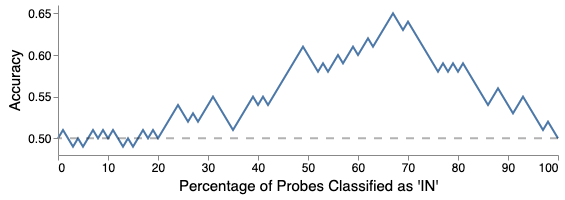}
    \caption{How the attack accuracy on Alice set changes when probe groups are sorted by Perceptron output score and the threshold to classify them as \textbf{in} is varied.
    }
    \label{fig:grouping_probes_threshold}
\end{figure}

\textcolor{black}{
Another possible approach to handle the problem of classifiers always predicting \textbf{in} is to consider the relative size of classifier output score. We can rank the samples by the classifier output scores, and decide top N\% to be \textbf{in} and rest to be \textbf{out}. Figure~\ref{fig:grouping_probes_threshold} shows how the accuracy changes when varying the \textbf{in} percentage. We can see that the accuracy can be much higher than the original result, especially if Bob can adjust the threshold based on his knowledge of \textbf{in} percentage in the probe.
}

\textcolor{black}{
This is the first strong general result for Bob, suggesting the membership inference attacks are possible if probes are defined as groups of sentences.\footnote{\textcolor{black}{We can imagine an alternative definition of this group-level membership inference where Bob's goal is to predict the percentage of overlap with respect to Alice's training data. This assumes that model trainers make corpus-level decisions about what data to train on. Reformulation of a binary problem to a regression problem may be useful for some purposes.}} Importantly, note that the classifier threshold adjustment is performed only for the classifiers in this section, and is not relevant for the classifiers in  Section~\ref{subsec:mainresult} to \ref{subsec:result_oov}.}




\subsection{Attacks using External Resources}
\label{subsec:externalresource}

\textcolor{black}{
Our results in Section \ref{subsec:mainresult} demonstrate the difficulty of general membership inference attacks. One natural question is whether attacks can be improved with even stronger features or classifiers, in particular by exploiting external resources beyond the dataset Carol provided to Bob.
We tried two different approaches: one using a Quality Estimation model trained on additional data, and another using a neural sequence model with a pre-trained language model.
}

\textcolor{black}{
Quality Estimation (QE) is a task of predicting the quality of a translation at the sentence or word level. One may imagine that a QE model might produce useful feature to tease apart \textbf{in} and \textbf{out} because \textbf{in} translations may have detectable improvements in quality. 
To train this model, we used the external dataset from the WMT shared task on QE~\cite{11372/LRT-2619}. Note that for our language pair, German to English, the shared task only had labeled dataset for SMT system. Our models are NMT, so the estimation quality may not be optimally matched, but we believe this is the best data available at this time. 
We applied the Predictor-Estimator~\cite{kim-etal-2017-predictor} implemented in an open source QE framework \texttt{OpenKiwi} ~\cite{kepler-etal-2019-openkiwi}. It consists of predictor that predicts each token of the target sentence given the target context and the source, and estimator that takes features produced by the predictor to estimate the labels; Both are made of LSTMs. We employed this model as this is one of the best models seen in the shared tasks, and it does not require alignment information.
The model metrics on the WMT18 dev set, namely Pearson's correlation, Mean Average Error and Root Mean Squared Error for sentence-level scores are 0.6238, 0.1276, and 0.1745 respectively.
}

\textcolor{black}{
We used the sentence score estimated by the QE model as an extra feature for classifiers described in Section~\ref{subsec:mainresult}. The results are shown in Table~\ref{table:result_external_resources}. We can see that this extra feature did not give any significant influence to the accuracy. 
In a more detailed analysis, we find that the reason is that our \textbf{in} and \textbf{out} probes both contain a range of translations from low to high quality translations, and our QE model may not be sufficiently fine-grained to tease apart any potential differences. In fact, this may be difficult even for a human estimator. 
}

\textcolor{black}{
Another approach to exploit external resources is to use language model pre-trained on a large amount of text. In particular, we used BERT~\cite{devlin-etal-2019-bert} which has shown competitive results in many NLP tasks. We used BERT directly as a classifier, and followed a fine-tuning setup similar to paraphrase detection: for our case the inputs are the English translation and reference sentences, and the output is the binary membership label. This setup is similar to the classifiers we described in Section \ref{subsec:attacks_classifier}, where rather than training Perceptron or Decision Tree on manually-defined features, we directly applied BERT-based sequence encoders on the raw sentences.
}

\textcolor{black}{
We fine-tuned the BERT \texttt{Base,Cased} English model with Bob:train. The results are shown in  Table~\ref{table:result_external_resources}. Similar to previous results, the accuracy is 50\% so the attack using BERT as classifier was not successful. 
Detailed examination of the BERT classifier probabilities show that they are scattered around 0.5 for all cases, but in general quite random for both Bob and Alice probes. This result is similar to the other simpler classifiers in Section \ref{subsec:mainresult}. 
}

\begin{table}[h]
\centering
\setlength\tabcolsep{5pt}
\begin{tabular}{ l | c : c c c } 
 \hline
  & Alice & Bob:train & Bob:valid & Bob:test  \\ 
 \hline
P & 50.0 & 49.9 & 50.0 & 50.0 \\
DT & 50.3 & 51.4 & 51.1 & 51.1 \\
NB & 50.4 & 51.2 & 51.1 & 51.0 \\
NN & 49.8 & 66.1 & 50.0 & 50.1 \\
MLP & 50.4 & 51.0 & 51.0 & 50.8 \\ \hdashline
BERT & 50.0 & 50.0 & 50.0 & 50.0 \\
 \hline
\end{tabular}
\caption{Membership inference accuracies for classifiers with Quality Estimation sentence score as an extra feature, and a BERT classifier.}
\label{table:result_external_resources}
\end{table}

\textcolor{black}{
In summary, from above results we can see that even with external resources and more complex classifiers, sentence-level attack is still very difficult for Bob. 
We believe this attests to the inherent difficulty of the sentence-level membership inference problem.
}

\section{Discussions and Conclusions}
\label{sec:discussion_conclusion}

We formalized the problem of membership inference attacks on sequence generation tasks, and used Machine Translation as an example to investigate the feasibility of a privacy attack.

Our results in Section \ref{subsec:mainresult}
\textcolor{black}{and Section \ref{subsec:externalresource}}
show that Alice is generally safe and it is difficult for Bob to infer the sentence-level membership. 
In contrast to attacks on \textit{standard classification} problems \cite{shokri2017membership}, \textit{sequence generation} problems maybe be harder to attack because the input and output spaces are far larger and complex, making it difficult to determine the quality of the model output or how confident the model is. Also, the output distribution of class labels is an effective feature for the attacker for standard classification problems, but is difficult to exploit in the sequence case.

However, this does not mean that Alice has no risk of leaking private information. Our analyses in Sections \ref{subsec:result_ood} and \ref{subsec:result_oov} show that Bob's accuracy on out-of-domain and out-of-vocabulary data is above chance, suggesting that attacks may be feasible in conditions where unseen words and domains cause the model to behave differently. 
Further, Section \ref{subsec:grouping} shows that for a looser definition of membership attack on groups of sentences, the attacker can win at a level above chance. 

Our attack approach was a simple one, using shadow models to mimic the target model. Bob can attempt more complex strategies, for example, by using the translation API multiple times per sentence.
Bob can manipulate a sentence, for example, by dropping or adding words, and observe how the translation changes. 
We may also use the metrics proposed by \newcite{carlini2018memorization} as features for Bob; they show how recurrent models might unintentionally memorize rare sequences in the training data, and proposed a method to detect it. 
Bob can also add ``watermark sentences'' that have some distinguishable characteristics to influence the Alice model, making attack easier.
To guard against these attack, Alice protection strategy may include random subsampling of training data or additional regularization terms.

Finally, we note some important caveats when interpreting our conclusions. 
The translation quality of Alice and Bob MT models turned out to be similar in terms of BLEU. This situation favors Bob, but in practice Bob is not guaranteed to be able to create shadow models of the same standard, nor verify how well it performs compared to the Alice model.
\textcolor{black}{
We stress that when one is to interpret the results, one must evaluate both on Bob's test set and Alice probes side-by-side, like those shown in Tables \ref{table:result_basic}, \ref{table:result_mt_score}, and \ref{table:result_external_resources}, to account for the fact that Bob's attack on his own shadow model translations is likely an optimistic upper-bound on the real attack accuracy on Alice's model.
}

We believe our dataset and analysis is a good starting point for research in these privacy questions. 
While we focused on MT, the formulation is applicable to other kinds of sequence generation models such as text summarization and video captioning; these will be interesting as future work.

\section*{Acknowledgments}

The authors thank the anonymous reviewers and the action editor, Colin Cherry, for their comments.

\bibliography{tacl2018}
\bibliographystyle{acl_natbib}

\end{document}